\def\year{2020}\relax
\documentclass[letterpaper]{article} 
\usepackage{aaai20}  
\usepackage{times}  
\usepackage{helvet} 
\usepackage{courier}  
\usepackage[hyphens]{url}  
\usepackage{graphicx} 
\usepackage{graphicx}  
\usepackage{url}
\usepackage{wrapfig}
\usepackage{amsmath}
\usepackage{amssymb}
\usepackage{color,xcolor,colortbl}
\usepackage{enumitem}
\usepackage{algorithm}
\usepackage{algorithmic}
\usepackage{bm,bbm}
\usepackage{booktabs}
\usepackage{mathtools}
\usepackage{array}
\usepackage{subcaption}
\usepackage{pbox}
\usepackage{pifont}
\usepackage{multirow}
\usepackage{tikz}
\usepackage{tikz-cd}
\usetikzlibrary{arrows,arrows.meta}
\tikzcdset{arrow style=tikz, diagrams={>=stealth'}}
\usepackage{stfloats}

\usepackage{pifont}
\newcommand{\done}{\rlap{$\square$}{\raisebox{0pt}{\large\hspace{-3.5pt} \textsf{x}}}%
\hspace{2pt}}

\newcommand\numberthis{\addtocounter{equation}{1}\tag{\theequation}}

\definecolor{darkblue}{rgb}{0.0, 0.0, 0.55}
\newenvironment{fontppl}{\fontfamily{ppl}\selectfont}{\par} 
\definecolor{input}{rgb}{0.63, 0.79, 0.95}
\definecolor{conv1}{rgb}{1.0, 0.89, 0.77}
\definecolor{maxpool}{rgb}{0.92, 0.3, 0.26}
\definecolor{dense}{rgb}{0.82, 0.62, 0.91}
\definecolor{relu}{rgb}{0.76, 0.33, 0.76}

\usepackage{multirow}
\usepackage{comment}
\usepackage{arydshln}

\title{Joint Parsing and Generation for Abstractive Summarization}
\author{
\textbf{Kaiqiang Song,\textsuperscript{\rm 1} Logan Lebanoff,\textsuperscript{\rm 1} Qipeng Guo\textsuperscript{\rm 2}}\\
\Large \textbf{Xipeng Qiu,\textsuperscript{\rm 2} Xiangyang Xue,\textsuperscript{\rm 2} Chen Li,\textsuperscript{\rm 3} Dong Yu,\textsuperscript{\rm 3} Fei Liu\textsuperscript{\rm 1}}\\
\textsuperscript{\rm 1}Computer Science Department, University of Central Florida\\
\textsuperscript{\rm 2}School of Computer Science, Fudan University\;
\textsuperscript{\rm 3}Tencent AI Lab, Bellevue, WA\\
\{kqsong,loganlebanoff\}@knights.ucf.edu, \{qpguo16,xpqiu,xyxue\}@fudan.edu.cn\\ \{ailabchenli,dyu\}@tencent.com, feiliu@cs.ucf.edu}

\begin{document}

\maketitle

\begin{abstract}

Sentences produced by abstractive summarization systems can be ungrammatical and fail to preserve the original meanings, despite being locally fluent. 
In this paper we propose to remedy this problem by jointly generating a sentence and its syntactic dependency parse while performing abstraction.
If generating a word can introduce an erroneous relation to the summary, the behavior must be discouraged. The proposed method thus holds promise for producing grammatical sentences and encouraging the summary to stay true-to-original. Our contributions of this work are twofold. First, we present a novel neural architecture for abstractive summarization that combines a sequential decoder with a tree-based decoder in a synchronized manner to generate a summary sentence and its syntactic parse. Secondly, we describe a novel human evaluation protocol to assess if, and to what extent, a summary remains true to its original meanings. We evaluate our method on a number of summarization datasets and demonstrate competitive results against strong baselines.

\end{abstract}

\section{Introduction}
\label{sec:intro}

It is crucial for a summary to not only condense the source text but also render itself grammatical.
Without grammatical sentences, a summary can be ineffective,
because human brain derives meaning from the sentence as a whole rather than individual words. 
Abstractive summarization has made considerable recent progress~\cite{See:2017,Chen:2018:ACL,Kryscinski:2018}.  
Nonetheless, studies suggest that system summaries remain imperfect. 
A summary sentence can be ungrammatical and fail to convey the intended meaning, despite its local fluency~\cite{Song:2018,lebanoff-etal-2019-analyzing}.
In Table~\ref{tab:example}, we show example abstractive summaries produced by neural abstractive summarizers.
The first summary has failed to conform to grammar and other summaries changed the original meanings. 
These summaries not only mislead the reader but also hinder the applicability of summarization techniques in real-world scenarios.

\begin{table}[t]
\setlength{\tabcolsep}{5pt}
\renewcommand{\arraystretch}{0.9}
\centering
\begin{scriptsize}
\begin{tabular}{ll}
\toprule
\textbf{Source} &
\textsf{Today, because of a CNN story and the generosity of}\\
& \textsf{donors from around the world, Kekula wears scrubs}\\
& \textsf{bearing the emblem of the Emory University ...}\\[0.5em]
\textbf{Summ.} & 
\textsf{\emph{\textcolor{red}{CNN story and generosity of donors from around the world,}}} \\
& \textsf{\emph{\textcolor{red}{Kekula wears scrubs ...}}}\\
\midrule
\textbf{Source} &
\textsf{In its propaganda, ISIS has been using Abu Ghraib and}\\
& \textsf{other cases of Western abuse to legitimize its current}\\
& \textsf{actions in Iraq as the latest episodes ...}\\[0.5em]
\textbf{Summ.} & 
\textsf{\emph{\textcolor{red}{In its propaganda, ISIS is being used by the Islamic State}}}\\
& \textsf{\emph{\textcolor{red}{in Iraq and Syria ...}}}\\
\midrule
\textbf{Source} &
\textsf{Both state and foreign investments in Vietnam's agriculture}\\
& \textsf{have been not sufficient enough, while local farmers have}\\
& \textsf{to pay fees to contribute to building rural roads ...
}\\[0.5em]
\textbf{Summ.} & 
\textsf{\emph{\textcolor{red}{Vietnam's agriculture not sufficient enough}}}\\
\bottomrule
\end{tabular}
\end{scriptsize}
\caption{Example summaries generated by neural abstractive summarizers.
They are manually re-cased for readability.}
\label{tab:example}
\end{table}

In this paper, we attempt to remedy this problem by introducing a new architecture to jointly generate a summary sentence and its syntactic parse, while performing abstraction. 
This is a non-trivial task, as the method must tightly couple summarization and parsing algorithms, which are two significant branches of NLP.
A joint model for generating summary sentences and parse trees can be more appealing than a pipeline method. 
The latter may suffer from error propagation, e.g., an ill-formed summary sentence can lead to more parsing errors.
Further, a joint method mimics the human behavior,
e.g., an editor writes a summary and makes corrections instantly as the text is written.
She needs not to finish the whole summary in order to correct errors. 
A method that incrementally produces a summary sentence and its syntactic parse aligns with this observation.

Our proposed joint model seeks to transform the \emph{source} sequence to a linearized parse tree of the \emph{summary} sequence.  
The model seamlessly integrates a shift-reduce dependency parser into a summarization system employing the encoder-decoder architecture.
A ``\textsc{shift}'' operation leads the summarizer to generate a new word by copying it from the source text or choosing a word from the vocabulary; 
whereas a ``\textsc{reduce}'' operation adds a dependency arc between words of the partial summary.
The challenge of this task is to construct effective representations that support both tasks, as they require different contextual representations.
We propose to couple a sequential decoder for predicting new summary words and a tree-based decoder for predicting dependency arcs, and ensure both decoders work in a synchronized fashion.
We also introduce an important addition making use of \emph{topological sorting} of tree nodes to accelerate the training procedure, making the framework computationally feasible.
Our research contributions can be summarized as follows:

\begin{itemize}[topsep=3pt,itemsep=-1pt,leftmargin=*]

\item we propose to simultaneously decode sentences and their syntactic parses while performing abstraction.
Our work represents a first attempt toward joint abstractive summarization and parsing that holds promise for improved sentence grammaticality and truthful summaries;

\item we present a novel neural architecture coupling a sequential and a tree decoder to generate summary sentences and parse trees simultaneously. 
Experiments are performed on a variety of summarization datasets to demonstrate the effectiveness of the proposed method;

\item we describe a new human evaluation protocol to assess if an abstractive summary has preserved the original meanings, and importantly, if it has introduced any new meanings that are nonexistent in the original text.
The last factor is largely under-investigated in the literature.\footnote{We make our implementation and models publicly available at\\ https://github.com/ucfnlp/joint-parse-n-summarize}

\end{itemize}

\section{Related Work}
\label{sec:related}

Recent years have seen increasing interest in summarization using encoder-decoder models~\cite{Rush:2015,Nallapati:2016,See:2017,Celikyilmaz:2018,lebanoff-etal-2019-scoring}.
An encoder condenses the source text to a fix-length vector and a decoder unrolls it to a summary. 
An encoder (or decoder) can be realized using recurrent networks~\cite{Chen:2016,Tan:2017,Cohan:2018,Lebanoff:2018,Gehrmann:2018}, convolutional networks~\cite{Chopra:2016,Narayan:2018:EMNLP}, or Transformer~\cite{Devlin:2018,Liu:2018:ICLR,Song:2020}.
To generate a summary word, a decoder can copy a word from the source text or select an unseen word from the vocabulary.
This flexibility allows for diverse lexical choices.
Nevertheless, with greater flexibility comes the increased risk of producing ill-formed summary sentences that are ungrammatical and fail to preserve the original meanings.

Parsing the source text to identify summary-worthy textual units has been exploited in the past.
Marcu~\shortcite{Marcu:1997,Marcu:1998} utilizes discourse structure generated by an RST parser to identify summary units that are central to the claims of the document.
A number of recent studies have explored constituency and dependency grammars~\cite{Daume:2002,Clarke:2008,Martins:2009,Filippova:2010,Kirkpatrick:2011,Wang:2013,Durrett:2016}, rhetorical structure~\cite{Christensen:2013,Yoshida:2014,Li:2016:EDU}, and abstract meaning representation~\cite{Liu:2015:NAACL,Liao:2018,Hardy:2018} to generate compressive and abstractive summaries.
In this paper we emphasize that \emph{target-side} syntactic analysis is especially important to ensure the well-formedness of abstractive summaries,
because generating summary words and predicting relations between words are interleaved operations.

Summarization and parsing are traditionally regarded as separate tasks.
These systems are now both realized using neural sequence-to-sequence models, making it possible to tackle both tasks in a single framework.  
There have been a variety of studies examining neural dependency parsers using transition- and graph-based algorithms~\cite{Dyer:2015,Kiperwasser:2016,Dozat:2017,Ma:2018:ACL}.
Our method, inspired by the recurrent neural network grammar (RNNG; Dyer et al., 2016)\nocite{Dyer:2016} that describes a generative probabilistic model for parsing and language modeling~\cite{Kuncoro:2017}, offers a way to perform summary generation and parsing in a synchronized manner.
Incorporating syntax is found to improve translation~\cite{Li:2017:NMT,eriguchi-etal-2017-learning,Wu:2017,Wang:2018:NMT}.
But to date, there has been little work to simultaneously generate a sentence and its syntactic parse, combining summarization with parsing techniques.
Our aim is not to improve existing parsers but to leveraging parsing for abstractive summarization.
Parsing is essentially a structured prediction problem, whereas summarization involves \emph{information reduction} from source to target, which poses an important challenge.
In the following section, we describe our model in detail.

\section{Our Approach}
\label{sec:our_approach}

Our goal is to transform a source text $\mathbf{x}$ containing one or more sentences to a target sequence containing a linearized parse tree of the summary, represented by $\mathbf{y}^\mathcal{T}$.
We expect a summary to contain a single sentence, as our focus is to improve sentence grammaticality.\footnote{When a multi-sentence summary is desired, it is possible to generate summary sentences repeatedly from selected subsets of source sentences, as suggested by recent studies~\cite{Chen:2018:ACL,Gehrmann:2018}.}
We use dependency grammar as syntactic representation of the summary.
Dependency is useful for semantic tasks and transition-based parsing algorithms are efficient, linear-time in the sequence length.\footnote{Our method is also general enough to allow other syntactic/semantic formalisms such as the constituency grammar or abstract meaning representation~\cite{Banarescu:2013,Konstas:2017} to be exploited in future work.}

\vspace{0.05in}
\noindent\textbf{Problem formulation} \quad
Our target sequence $\mathbf{y}^\mathcal{T}$ consists of interleaved \textsc{gen}($w$) and \textsc{reduce-l/r} operations that incrementally build a dependency parse tree. 
Table~\ref{tab:rnng} shows an example.
The second column contains $\mathbf{y}^\mathcal{T}$ and the third column contains partial dependency trees stored in a \emph{stack}.
A \textsc{gen}($w$) operation pushes a summary word $w$ to the stack;
\textsc{reduce-l} creates a left arc between the top and second top word in the stack, where the top word is the head; \textsc{reduce-r} creates a right arc where the top word is the dependent. 
We choose not to label the arcs, as this work focuses on generating well-structured sentences but not on predicting labels.
The decoding process comes to an end when there is a single tree remaining in the stack. 
A summary $\mathbf{y}$ can be obtained from $\mathbf{y}^\mathcal{T}$ by retrieving all \textsc{gen} operations.

\begin{table}[t]
\setlength{\tabcolsep}{3.5pt}
\renewcommand{\arraystretch}{1.1}
\centering
\begin{footnotesize}
\begin{tabular}{c|l|l}
$t$ & $\mathbf{y}^\mathcal{T}$ & \textbf{Stack} \\
\toprule
1 & 
-- &
\begin{tikzcd}[row sep=0em, column sep=0.8em, math mode=false, /tikz/cells={nodes={inner xsep=0.2ex,inner ysep=0.2ex}}]
\textsf{R} (root node)
\end{tikzcd}\\
2 & 
\textsc{gen}(\emph{a}) &
\begin{tikzcd}[row sep=0em, column sep=0.8em, math mode=false, /tikz/cells={nodes={inner xsep=0.2ex,inner ysep=0.2ex}}]
\textsf{R} & \emph{a} 
\end{tikzcd}\\
3 & 
\textsc{gen}(\emph{man}) &
\begin{tikzcd}[row sep=0em, column sep=0.8em, math mode=false, /tikz/cells={nodes={inner xsep=0.2ex,inner ysep=0.2ex}}]
\textsf{R} & \emph{a} & \emph{man} 
\end{tikzcd}\\
4 & 
\textsc{reduce-l} &
\begin{tikzcd}[row sep=0em, column sep=0.8em, math mode=false, /tikz/cells={nodes={inner xsep=0.2ex,inner ysep=0.2ex}}]
\textsf{R} & \emph{a} & \emph{man} \arrow[l] 
\end{tikzcd}\\
5 & 
\textsc{gen}(\emph{escaped}) &
\begin{tikzcd}[row sep=0em, column sep=0.8em, math mode=false, /tikz/cells={nodes={inner xsep=0.2ex,inner ysep=0.2ex}}]
\textsf{R} & \emph{a} & \emph{man} \arrow[l] & \emph{escaped} 
\end{tikzcd}\\
6 & 
\textsc{reduce-l} &
\begin{tikzcd}[row sep=0em, column sep=0.8em, math mode=false, /tikz/cells={nodes={inner xsep=0.2ex,inner ysep=0.2ex}}]
\textsf{R} & \emph{a} & \emph{man} \arrow[l] & \emph{escaped} \arrow[l]
\end{tikzcd}\\
{\cellcolor[gray]{.8}}7 & {\cellcolor[gray]{.8}}\textsc{gen}(\emph{from}) 
& {\cellcolor[gray]{.8}}\begin{tikzcd}[row sep=0em, column sep=0.8em, math mode=false, /tikz/cells={nodes={inner xsep=0.2ex,inner ysep=0.2ex}}]
\textsf{R} & \emph{a} & \emph{man} \arrow[l] & \emph{escaped} \arrow[l] & \emph{from} 
\end{tikzcd}\\
8 & 
\textsc{gen}(\emph{prison}) &
\begin{tikzcd}[row sep=0em, column sep=0.8em, math mode=false, /tikz/cells={nodes={inner xsep=0.2ex,inner ysep=0.2ex}}]
\textsf{R} & \emph{a} & \emph{man} \arrow[l] & \emph{escaped} \arrow[l] & \emph{from} & \emph{prison}
\end{tikzcd}\\
9 & 
\textsc{reduce-l} &
\begin{tikzcd}[row sep=0em, column sep=0.8em, math mode=false, /tikz/cells={nodes={inner xsep=0.2ex,inner ysep=0.2ex}}]
\textsf{R} & \emph{a} & \emph{man} \arrow[l] & \emph{escaped} \arrow[l] & \emph{from} & \emph{prison} \arrow[l]
\end{tikzcd}\\
10 & 
\textsc{reduce-r} &
\begin{tikzcd}[row sep=0em, column sep=0.8em, math mode=false, /tikz/cells={nodes={inner xsep=0.2ex,inner ysep=0.2ex}}]
\textsf{R} & \emph{a} & \emph{man} \arrow[l] & \emph{escaped} \arrow[l] \arrow[rr, bend left=15] & \emph{from} & \emph{prison} \arrow[l]
\end{tikzcd}\\
11 & 
\textsc{reduce-r} &
\begin{tikzcd}[row sep=0em, column sep=0.8em, math mode=false, /tikz/cells={nodes={inner xsep=0.2ex,inner ysep=0.2ex}}]
\textsf{R} \arrow[rrr, bend left=15] & \emph{a} & \emph{man} \arrow[l] & \emph{escaped} \arrow[l] \arrow[rr, bend left=15] & \emph{from} & \emph{prison} \arrow[l]
\end{tikzcd}\\
\bottomrule
\end{tabular}
\end{footnotesize}
\caption{Illustration of the decoding process.
A summary sentence ``\emph{a man escaped from prison}'' and its dependency structure are simultaneously generated. 
The second column shows the target sequence $\mathbf{y}^\mathcal{T}$ and the third column contains partial parse trees stored in a \emph{stack}.
}
\label{tab:rnng}
\vspace{-0.1in}
\end{table}

We aim to predict the target sequence $\mathbf{y}^\mathcal{T}$ conditioned on the source $\mathbf{x}$.
The process proceeds incrementally.
As illustrated in Eq.~(\ref{eq:p_y_t}), $P(\mathbf{y}^\mathcal{T}|\mathbf{x})$ is factorized over time steps.
$P(\mathbf{y}^\mathcal{T}_{t} = o | \mathbf{y}^\mathcal{T}_{<t}, \mathbf{x})$ denotes the probability of a parsing operation, where $o \in$ \{\textsc{reduce-l}, \textsc{reduce-r}, \textsc{gen}\} and \textsc{gen} is unlexicalized.
$P(\mathbf{y}^\mathcal{T}_{t}=w | \mathbf{y}^\mathcal{T}_{<t}, \mathbf{x})$ represents the probability of generating a summary word $w$ at the $t$-th step;
the word can either be copied from the source text or selected from the vocabulary.
\begin{align*}
P(\mathbf{y}^\mathcal{T}|\mathbf{x}) = \displaystyle\prod_{t} 
& \Big[\underbrace{P(\mathbf{y}^\mathcal{T}_t = o|\mathbf{y}^\mathcal{T}_{<t}, \mathbf{x})}_{\mbox{\footnotesize\fontppl parsing}} 
\numberthis\label{eq:p_y_t}\\
\times & \underbrace{P(\mathbf{y}^\mathcal{T}_t = w|\mathbf{y}^\mathcal{T}_{<t}, \mathbf{x})^{\mathbbm{1}[o = \textsc{gen}]}}_{\mbox{\footnotesize\fontppl summarization}}\Big]
\end{align*}

At training time, the ground-truth sequence $\hat{\mathbf{y}}^\mathcal{T}$ is available, $P(\hat{\mathbf{y}}^\mathcal{T}_t = w |\hat{\mathbf{y}}^\mathcal{T}_{<t}, \mathbf{x})$ needs only be computed for certain steps where the parsing operation is \textsc{gen}, as indicated by $\mathbbm{1}[o = \textsc{gen}]$.
Our loss term corresponds to the conditional log-likelihood which can be separately calculated for parsing and summarization operations (Eq.~(\ref{eq:log_p_y_t})). 
During inference, we calculate $P(\mathbf{y}^\mathcal{T}_t | \mathbf{y}^\mathcal{T}_{<t}, \mathbf{x})$ as a joint distribution over parsing and summarization operations, where $\mathbf{y}^\mathcal{T}_t \in$ \{\textsc{reduce-l}, \textsc{reduce-r}, \textsc{gen}($w$)\}.
{\medmuskip=1mu
\thinmuskip=1mu
\thickmuskip=1mu
\nulldelimiterspace=0pt
\scriptspace=0pt
\begin{align*}
\log P(\hat{\mathbf{y}}^\mathcal{T}| \mathbf{x}) = &\Big[\displaystyle\sum_{t} \log P(\hat{\mathbf{y}}^\mathcal{T}_t = o|\hat{\mathbf{y}}^\mathcal{T}_{<t}, \mathbf{x})\Big]
\numberthis\label{eq:log_p_y_t}\\
+ &\Big[\displaystyle\sum_{t:o_t = \textsc{gen}} \log P(\hat{\mathbf{y}}^\mathcal{T}_t = w |\hat{\mathbf{y}}^\mathcal{T}_{<t}, \mathbf{x})\Big]
\end{align*}}

\begin{figure}
\centering
\includegraphics[width=3.05in]{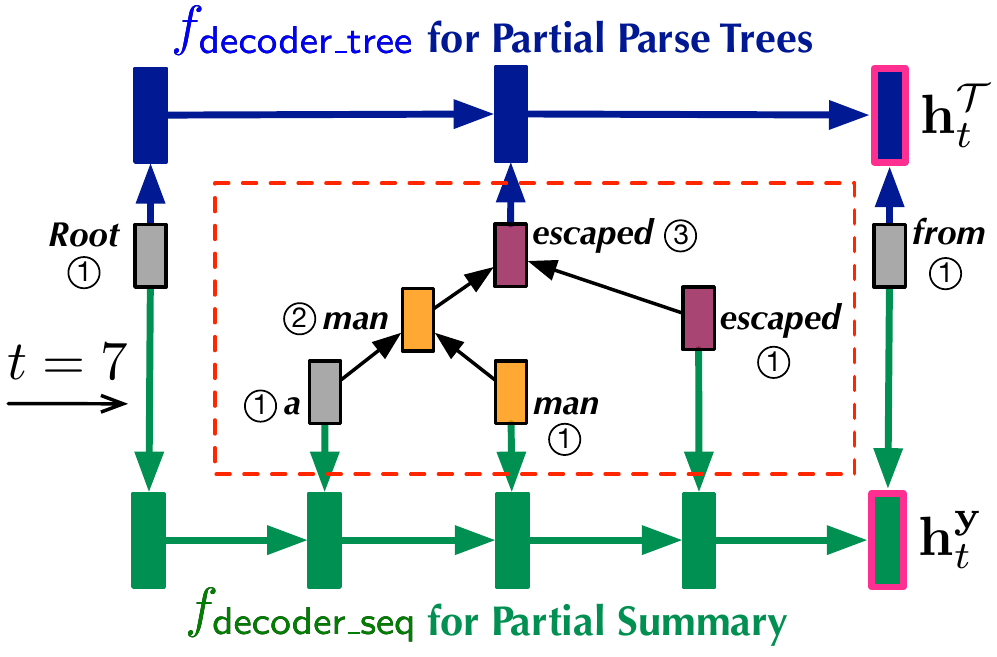}
\captionof{figure}{
$f_{\mbox{\scriptsize \textsf{decoder\_tree}}}$ (top) consumes the \emph{partial tree} representations of time $t$ one by one to build hidden representation $\mathbf{h}^\mathcal{T}_{t}$;
$f_{\mbox{\scriptsize \textsf{decoder\_seq}}}$ (bottom) consumes the embeddings of summary words to build \emph{partial summary} representation $\mathbf{h}_{t}^\mathbf{y}$.
}
\label{fig:rnng}
\end{figure}

\vspace{-0.05in}
\noindent\textbf{Neural representations}\quad
A crucial next step is to build neural representations to support both tasks.
Predicting the next parsing operation requires us to build an effective representation for \emph{partial parse trees}, denoted by $\mathbf{h}^\mathcal{T}_{t}$ at the $t$-th step, whereas predicting the next summary word suggests an effective representation for the \emph{partial summary}, represented by $\mathbf{h}^\mathbf{y}_t$.
We envision both tasks to benefit from a context vector $\mathbf{c}^\mathbf{x}_t$ that encodes source content that is deemed important for the $t$-th decoding step. 
We next describe a new architecture building representations for $\mathbf{h}^\mathcal{T}_{t}$, $\mathbf{h}^\mathbf{y}_t$, and $\mathbf{c}^\mathbf{x}_t$.

We model partial trees using stack-LSTM~\cite{Dyer:2015,Dyer:2016}.
Our stack maintains a set of partial trees at any time $t$; they are shown in the $t$-th row of Table~\ref{tab:rnng}.
For each partial tree, we build a vector representation for it by recursively applying a syntactic composition function (Eq.~(\ref{eq:g_head})).
The representation is built from bottom up, shown in the dotted circle of Figure~\ref{fig:rnng}.
A left arc (\textsc{reduce-l}) pops two elements from the stack.
It then applies the composition function to create a new representation $\mathbf{g}_{\mbox{\scriptsize \textsf{new\_head}}}$ and push it onto the stack;
similarly for right arc (\textsc{reduce-r}).
A \textsc{gen}($w$) operation pushes the embedding of a summary word $\mathbf{e}(w)$ to the stack.\footnote{$\mathbf{e}(w)$ has the same size as partial tree representations $\mathbf{g}$.}
Kuncoro et al.~\shortcite{Kuncoro:2017} report that the composition function learns to compute a tree representation by preserving the semantics of the head word, which fits our task.
{\medmuskip=1mu
\thinmuskip=1mu
\thickmuskip=1mu
\nulldelimiterspace=0pt
\scriptspace=0pt
\begin{align*}
& \mathbf{g}_{\mbox{\scriptsize \textsf{new\_head}}} = \tanh(\mathbf{W}^g[\mathbf{g}_{\mbox{\scriptsize \textsf{head}}} || \mathbf{g}_{\mbox{\scriptsize \textsf{dependent}}}] + \mathbf{b}^g)
\numberthis\label{eq:g_head}
\end{align*}}
\vspace{-0.18in}

We introduce an LSTM, denoted by $f_{\mbox{\scriptsize \textsf{decoder\_tree}}}$, to consume the \emph{partial tree} representations of time $t$ one by one to build the hidden representation $\mathbf{h}^\mathcal{T}_{t}$.
An illustration is presented in Figure~\ref{fig:rnng}.
E.g., when $t$=7, the stack contains 3 partial trees and we build a vector representation for each.
$f_{\mbox{\scriptsize \textsf{decoder\_tree}}}$ is unrolled 3 steps and its last hidden state is $\mathbf{h}^\mathcal{T}_{t}$. 
Similarly, we build the \emph{partial summary} representation $\mathbf{h}_{t}^\mathbf{y}$ using an LSTM denoted by $f_{\mbox{\scriptsize \textsf{decoder\_seq}}}$, which consumes the embeddings of summary words.
For example, when $t$=7, there are 5 words in the partial summary. $f_{\mbox{\scriptsize \textsf{decoder\_seq}}}$ is  unrolled 5 steps and its last hidden state is used as $\mathbf{h}^\mathbf{y}_{t}$. 
Note that for some steps, e.g., $t$=9, no summary words are generated, we copy $\mathbf{h}_{t}^\mathbf{y}$ from its previous step $\mathbf{h}_{t-1}^\mathbf{y}$.

A context vector ($\mathbf{c}_t^{\mathbf{x}}$) encoding the source content that is deemed important for the $t$-th decoding step is crucial to our method.
Important source content can not only aid in the prediction of future summary words but also parsing operations. 
We build the context vector $\mathbf{c}_t^\mathbf{x}$ in two steps.
First, we encode the source text $\mathbf{x}$ using a two-layer bidirectional LSTM denoted by $f_{\mbox{\scriptsize \textsf{encoder}}}$.
We use $\{\mathbf{h}_i^\mathbf{x}\}$ to denote the encoder hidden states, where $i$ is the index of source words. 
Next, we characterize the interaction between encoder and decoder hidden states using an attention mechanism (Eq.~(\ref{eq:s_t_i})).
We concatenate the partial tree and partial summary representations $[\mathbf{h}^\mathcal{T}_{t} || \mathbf{h}^\mathbf{y}_t]$ to form the decoder state.
The score $S_{t,i}$ measures the importance of the $i$-th source word to the $t$-th decoding step.
A context vector $\mathbf{c}_t^{\mathbf{x}}$ is then constructed as the weighted sum of source representations (Eq.~(\ref{eq:c_t})).
\begin{align*}
S_{t,i} &= \mathbf{w}^\top \tanh(\mathbf{W}^d [\mathbf{h}^\mathcal{T}_{t} || \mathbf{h}^\mathbf{y}_t] + \mathbf{W}^e \mathbf{h}_i^\mathbf{x})
\numberthis\label{eq:s_t_i}\\
\mathbf{c}_t^{\mathbf{x}} &= \mbox{softmax}(\mathbf{S}_{t}) \mathbf{h}^\mathbf{x}
\numberthis\label{eq:c_t}
\end{align*}

\noindent\textbf{Prediction}\,\,
We predict summary words $P(\mathbf{y}_{t}^\mathcal{T} = w |\mathbf{y}_{<t}^\mathcal{T}, \mathbf{x})$ and parsing operations $P(\mathbf{y}_{t}^\mathcal{T} = o |\mathbf{y}_{<t}^\mathcal{T}, \mathbf{x})$ with these representations.
We expect historical parsing operations to be helpful for the latter task, i.e., the sequence of \{\textsc{reduce-l}, \textsc{reduce-r}, \textsc{gen}($w$)\} operations shown in Table~\ref{tab:rnng}.
We thus use an LSTM to encode the sequence of past operations and its last hidden state is denoted by $\mathbf{h}_{t}^\mathcal{O}$.
A parsing operation is predicted based on $[\mathbf{h}_{t}^\mathcal{T} || \mathbf{h}_{t}^\mathcal{O}|| \mathbf{c}_t^\mathbf{x}]$, and we apply the softmax to obtain a distribution over parsing operations (Eq.~(\ref{eq:y_t_o})).
\begin{align*}
&\widetilde{\mathbf{h}}_t^\mathcal{T} = \tanh(\mathbf{W}^a [\mathbf{h}_{t}^\mathcal{T} || \mathbf{h}_{t}^\mathcal{O}|| \mathbf{c}_t^\mathbf{x}] + \mathbf{b}^a)
\numberthis\label{eq:h_t_t}\\
&P(\mathbf{y}_t^\mathcal{T} = o| \mathbf{y}_{<t}^\mathcal{T}, \mathbf{x}) = \mbox{softmax} (\mathbf{W}^o \widetilde{\mathbf{h}}_t^\mathcal{T})
\numberthis\label{eq:y_t_o}
\end{align*}

A summarizer should allow a summary word to be copied from the source text or generated from the vocabulary.
We implement a soft switch following See et al.~\shortcite{See:2017}, where $\lambda = \sigma(\mathbf{W}^z [\mathbf{h}_t^\mathbf{y} || \mathbf{h}_t^\mathcal{T} || \mathbf{c}_t^\mathbf{x}]) + b^z)$ is the likelihood of generating a summary word from the vocabulary.
The generation probability is defined in Eqs.~(\ref{eq:h_t_y_tilde}-\ref{eq:p_y_t_tilde}).
If a word $w$ occurs once or more times in the source text, its copy probability ($\sum_{i:w_i = w} \alpha_{t,i}$) is the sum of its attention scores over all the occurrences, where $\alpha_{t,i}$=$\mbox{softmax}_i(\mathbf{S}_t)$.
If a word $w$ appears in both the vocabulary and source text, $P(\mathbf{y}_t^\mathcal{T} = w | \boldsymbol\cdot)$ is a weighted sum of the generation and copy probabilities.
{\medmuskip=1mu
\thinmuskip=1mu
\thickmuskip=1mu
\nulldelimiterspace=0pt
\scriptspace=0pt
\begin{align*}
&\widetilde{\mathbf{h}}_t^\mathbf{y} = \tanh(\mathbf{W}^c [\mathbf{h}_{t}^\mathbf{y} || \mathbf{h}_{t}^\mathcal{T} || \mathbf{c}_t^\mathbf{x}] + \mathbf{b}^c) 
\numberthis\label{eq:h_t_y_tilde}\\
&\widetilde{P}(\mathbf{y}_t^\mathcal{T} = w|\mathbf{y}_{<t}^\mathcal{T}, \mathbf{x}) \;= \mbox{softmax} (\mathbf{W}^w \widetilde{\mathbf{h}}_t^\mathbf{y})
\numberthis\label{eq:p_y_t_tilde}\\
&P(\mathbf{y}_t^\mathcal{T} = w | \boldsymbol{\cdot}) = \lambda \widetilde{P}(\mathbf{y}_t^\mathcal{T} = w | \boldsymbol{\cdot}) + (1 - \lambda) \textstyle\sum_{i:w_i = w} \alpha_{t,i}
\end{align*}}
\vspace{-0.1in}

\noindent\textbf{Acceleration}\quad
Obtaining partial tree representations ($\mathbf{h}_t^\mathcal{T}$) can be computationally expensive, because $\mathbf{h}_t^\mathcal{T}$ has to be computed bottom-up according to the topology of a parse tree.
Further, parse trees in a mini-batch exhibit distinct topology, making it difficult to execute parallely;  frameworks such as DyNet~\cite{Neubig:2017} often process one instance at a time.
In this work we instead propose to arrange the tree nodes of all instances into groups according to their topological order; representations for nodes of the same group ($\mathbf{h}_t^\mathcal{T}$) are computed in parallel. 
For example, in Figure~\ref{fig:rnng}, the nodes marked with ``1'' are first processed, followed by nodes marked with ``2'' and so forth.
This strategy allows for mini-batch training with parse trees of distinct topology and maximizing the usage of computing resources.

\section{Experiments}
\label{sec:experiments}

We present our datasets, settings, baselines, qualitative and quantitative evaluation of our proposed method. 
We then discuss our findings and shed light on future work.

\subsection{Data and Hyperparameters}
\label{sec:data}

We conduct experiments on a variety of datasets to gauge the effectiveness of our proposed method.
We experiment with \textsc{Gigaword}~\cite{Parker:2011} and \textsc{Newsroom}~\cite{Grusky:2018}.
\textsc{Gigaword} contains about 10M articles gathered from seven news sources (1995-2010); 
\textsc{Newsroom} is a more recent effort containing 1.3M articles (1998-2017) collected from 38 news agencies.
We use the standard data splits and follow the same procedure as Rush et al.~\shortcite{Rush:2015} to process both datasets. 
The task of \textsc{Gigaword} and \textsc{Newsroom} is to reduce the first sentence of a news article to a title-like summary. 

The CNN/DM dataset~\cite{Hermann:2015} has been extensively studied. 
We use the version provided by See et al.~\shortcite{See:2017} but formulate it as a sentence summarization task.
We aim to condense a source sentence to a well-formed summary sentence.
The source sentences are  obtained by pairing each summary sentence with its most similar sentence in the article according to averaged R-1, R-2, and R-L F-scores~\cite{Lin:2004}.
We denote this \emph{reduced} dataset as ``\textsc{cnn/dm-r}.'' 
It is distinct from \textsc{Gigaword} and \textsc{Newsroom} because its ground-truth summaries are full grammatical sentences, whereas the latter are article titles that appear enticing but not necessarily be full sentences.

\begin{table}
\setlength{\tabcolsep}{6.5pt}
\renewcommand{\arraystretch}{1.1}
\centering
\begin{small}
\begin{tabular}{lrrrr}
& \multicolumn{1}{c}{$|\mathbf{y}|$} & \multicolumn{1}{c}{\textbf{Train}} & \multicolumn{1}{c}{\textbf{Dev}} & \multicolumn{1}{c}{\textbf{Test}}\\
\toprule
\textsc{Gigaword} & 8.41 & 4,020,581 & 4,096 & 1,951\\ 
\textsc{Newsroom} & 10.18 & 199,341 & 21,530 & 21,382\\ 
\textsc{cnn/dm-r} & 13.89 & 472,872 & 25,326 & 20,122\\ 
\textsc{WebMerge} & 31.43 & 1,331,515 & 40,879 & 43,934\\ 
\bottomrule
\end{tabular}
\end{small}
\caption{Statistics of our datasets. $|\mathbf{y}|$ is number of words.
}
\label{tab:stats}
\end{table}

We further experiment on many-to-one sentence summarization, where the goal is to fuse multiple source sentences to a summary sentence.
Existing datasets for sentence fusion are often small, containing thousands of instances~\cite{Thadani:2013:IJCNLP}. 
In this work we present a novel use of a newly released dataset---WebSplit~\cite{Narayan:2017:EMNLP}. 
The dataset was originally developed for sentence \emph{simplification}, where a lengthy source sentence is to be converted to multiple, simpler sentences for ease of understanding.
Importantly, we swap the source and target sequences, so that the task becomes fusing multiple source sentences to a well-formed summary sentence. 
We name this task \textsc{WebMerge} to avoid confusion. 
On average, a source text contains 4.4 sentences and the target is a single sentence.
A (source, target) pair is accompanied by a set of semantic triples in the form of ``subject$|$property$|$object'' and the semantics remain unchanged during merging.  
We utilize these triples for human evaluation (\S\ref{sec:results}).
In Table~\ref{tab:stats}, we provide statistics of all datasets used in this study.

\vspace{0.05in}
\noindent\textbf{Hyperparameters}\quad
We create an input vocabulary to contains word appearing 5 times or more in the dataset;
the output vocabulary contains the most frequent 10k words.
We set all LSTM hidden states to be 256 dimensions.
Because datasets containing both summaries and human-annotated dependency parses are unavailable,
we use the Stanford parser~\cite{Chen:2014:EMNLP} to obtain parse trees for reference summaries.
During training, we use a batch size of 64 and Adam~\cite{Kingma:2015} for parameter optimization, with lr=1e-3, betas=[0.9,0.999], and eps=1e-8.
We apply gradient clipping of [-5,5], and a weight decay of 1e-6.
At decoding time, we apply beam search with reference~\cite{Tan:2017} to generate summary sequences.
$K$=10 is the beam size.

\begin{table}[t]
\setlength{\tabcolsep}{10pt}
\renewcommand{\arraystretch}{1.08}
\centering
\begin{small}
\begin{tabular}{lccc}
& \multicolumn{3}{c}{\textbf{Gigaword Test Set}}\\
\textbf{System} & \textbf{R-1} & \textbf{R-2} & \textbf{R-L} \\
\toprule
ABS & 29.55 & {11.32} & 26.42 \\
ABS+ & 29.76 & {11.88} & 26.96 \\
Luong-NMT & 33.10 & 14.45 & 30.71 \\
RAS-LSTM & 32.55 & {14.70} & 30.03 \\
RAS-Elman & 33.78 & {15.97} & 31.15 \\
ASC+FSC1 & 34.17 & {15.94} & 31.92 \\
lvt2k-1sent & 32.67 & {15.59} & 30.64 \\
lvt5k-1sent& 35.30 & {16.64} & 32.62 \\
Multi-Task & 32.75 & {15.35} & 30.82 \\
SEASS & 36.15 & 17.54 & 33.63 \\
DRGD & 36.27 & {17.57} & 33.62 \\
Struct+2Way+Word & 35.47 & {17.66} & {33.52}\\
EntailGen+QuesGen & 35.98 & {17.76} & {33.63}\\
\midrule
GenParse-\textsc{Base} (This work) & 35.21 & 17.10 & 32.88\\
GenParse-\textsc{Full} (This work) & \textbf{36.61} & \textbf{18.85} & \textbf{34.33} \\ 
\bottomrule
\end{tabular}
\end{small}
\caption{
Summarization results on Gigaword dataset.
Our GenParse systems perform on par with or superior to state-of-the-art systems on the standard test set. }
\label{tab:results_giga}
\end{table}

\begin{table*}
\setlength{\tabcolsep}{4pt}
\renewcommand{\arraystretch}{1.08}
\centering
\begin{small}
\begin{tabular}{ll|ccc|ccc|ccc}
& & \multicolumn{3}{c|}{\textbf{ROUGE-1}} & \multicolumn{3}{c|}{\textbf{ROUGE-2}} & \multicolumn{3}{c}{\textbf{ROUGE-L}}\\
& \textbf{System} & \textbf{P} & \textbf{R} & \textbf{F} & \textbf{P} & \textbf{R} & \textbf{F} & \textbf{P} & \textbf{R} & \textbf{F}\\
\toprule
\multirow{3}{*}{\textsc{Newsroom}} &
PointerGen{\scriptsize~\cite{See:2017}} & 43.73 & 38.83 & 39.94 & 21.82 & 18.97 & 19.56 & 40.15 & 35.65 & 36.66 \\
& GenParse-\textsc{Base} (This work) & 41.88 & 36.00 & 37.65 & 20.04 & 16.90 & 17.70 & 38.73 & 33.33 & 34.84 \\
& GenParse-\textsc{Full} (This work) & \textbf{45.17} & \textbf{39.77} & \textbf{41.06} & \textbf{23.48} & \textbf{20.17} & \textbf{20.89} & \textbf{41.82} & \textbf{36.81} & \textbf{38.01} \\
\midrule
\multirow{3}{*}{\textsc{cnn/dm-r}} &
PointerGen{\scriptsize~\cite{See:2017}} & {50.91} & 49.82 & 49.26 & {34.73} & 33.32 & 33.16 & {48.10} & 46.95 & 46.49 \\
& GenParse-\textsc{Base} (This work) & 48.24 & 46.52 & 46.46 & 31.44 & 29.62 & 29.82 & 45.43 & 43.72 &	43.71 \\
& GenParse-\textsc{Full} (This work) & 50.15 & \textbf{53.11} & \textbf{50.49} & 34.51 & \textbf{35.99} & \textbf{34.38} & 47.33 & \textbf{50.00} & \textbf{47.60} \\
\midrule
\multirow{3}{*}{\textsc{WebMerge}} & 
PointerGen{\scriptsize~\cite{See:2017}} & 54.73 & 49.22 & 50.90 & 25.80 & 23.08 & 23.89 & 40.60 & 36.67 & 37.84 \\
& GenParse-\textsc{Base} (This work) & 37.79 & 35.86 & 36.23 & 12.63 & 11.99 & 12.09 & 28.87 & 27.59 & 27.77\\
& GenParse-\textsc{Full} (This work) & \textbf{62.26} & \textbf{54.69} & \textbf{57.24} & \textbf{32.10} & \textbf{28.41} & \textbf{29.58} & \textbf{48.13} & \textbf{42.54} & \textbf{44.41} \\
\bottomrule
\end{tabular}
\end{small}
\caption{Summarization results on {Newsroom}, {CNN/DM-R}, and {WebMerge} datasets. Our GenParse-\textsc{Full} method jointly decodes a summary and its dependency structure using a novel architecture that performs competitively against strong baselines.
It outperforms both pointer-generator networks and the ablated model GenParse-\textsc{Base} without using the tree-decoder.
}
\label{tab:results_all}
\end{table*}

\subsection{Experimental Results}
\label{sec:results}

\noindent\textbf{Summarization}\quad
We present summarization results on all datasets.
Evaluation is performed using the automatic metric of ROUGE~\cite{Lin:2004}, 
which measures the n-gram overlap between system and reference summaries,
as well as human evaluation of grammaticality and preservation of meanings.
We discuss our findings at the end.

In Table~\ref{tab:results_giga}, we present summarization results on the Gigaword test set containing 1951 instances.
We are able to compare our system, denoted by \emph{GenParse}, with a variety of state-of-the-art neural abstractive summarizers; they are described below.
Our system can be a valuable addition to existing neural summarizers, as it performs summarization and parsing jointly on the target-side to improve sentence grammaticality.
We explore two variants of our system:
GenParse-\textsc{Full} represents the full model;
GenParse-\textsc{Base} is an ablated model where we drop the tree-decoder to test its impact on summarization performance; this corresponds to removing $\mathbf{h}_t^\mathcal{T}$ and $\mathbf{h}_{t}^\mathcal{O}$ in all equations.
All other components remain the same. 
As shown in Table~\ref{tab:results_giga}, 
our GenParse system performs on par with or superior to state-of-the-art systems on the standard Gigaword test set. 
The full model yields the highest R-2 score of 18.85.
It outperforms the GenParse-\textsc{Base} model, demonstrating the effectiveness of coupling a sequential decoder with a tree-based decoder in a synchronized manner.

\vspace{0.05in}
\begin{itemize}[topsep=3pt,itemsep=-1pt,leftmargin=*]
\begin{small}

\item\emph{ABS} and \emph{ABS+}~\cite{Rush:2015} are the first work using an encoder-decoder architecture for summarization;

\item\emph{Luong-NMT}~\cite{Chopra:2016} re-implements the attentive encoder-decoder of Luong et al.~\shortcite{Luong:2015};

\item\emph{RAS-LSTM} and \emph{RAS-Elman}~\cite{Chopra:2016} describe a convolutional attentive encoder that ensures the decoder focuses on appropriate words at each step of generation;

\item\emph{ASC+FSC1}~\cite{Miao:2016} presents a generative auto-encoding sentence compression model jointly trained on labelled/unlabelled data;

\item\emph{lvt2k-1sent} and \emph{lvt5k-1sent}~\cite{Nallapati:2016} address issues in the encoder-decoder model, including modeling keywords, capturing sentence-to-word structure, and handling rare words;

\item\emph{Multi-Task w/ Entailment}~\cite{Pasunuru:2018} combines entailment with summarization in a multi-task setting;

\item\emph{DRGD}~\cite{Li:2017:DRGD} describes a deep recurrent generative decoder learning latent structure of summary sequences via variational inference;

\item\emph{Struct+2Way+Word}~\cite{Song:2018} describes a structure infused copy mechanism for sentence summarization;

\item\emph{EntailGen+QuesGen}~\cite{Guo:2018:ACL} is a multi-task architecture to perform summarization with question generation and entailment generation in one framework.

\end{small}
\end{itemize}

In Table~\ref{tab:results_all} we present summarization results on the \textsc{Newsroom}, \textsc{cnn/dm-r}, and \textsc{WebMerge} datasets.
The task of \textsc{WebMerge} is to fuse multiple source sentences to a well-formed summary sentence while keeping the semantics unchanged;
the task of \textsc{Newsroom} and \textsc{cnn/dm-r} is sentence summarization, but not document summarization.
Because of that, the ROUGE scores presented in Table~\ref{tab:results_all} should not be directly compared with other published results.
Instead, we train the pointer-generator networks with coverage mechanism (PointerGen; See et al. 2017)\nocite{See:2017}, one of the best performed neural abstractive summarizers, on the train split of each dataset, then report results on the test split;
we apply a similar process to our GenParse systems.
We observe that the GenParse-\textsc{Full} model consistently outperforms strong baselines across all datasets.
The results are outstanding because our system jointly performs summarization and dependency parsing;
it involves an increased task complexity than performing summarization only; and our full model is able to excel on this task.

\begin{figure}[t]
\centering
\includegraphics[width=3in]{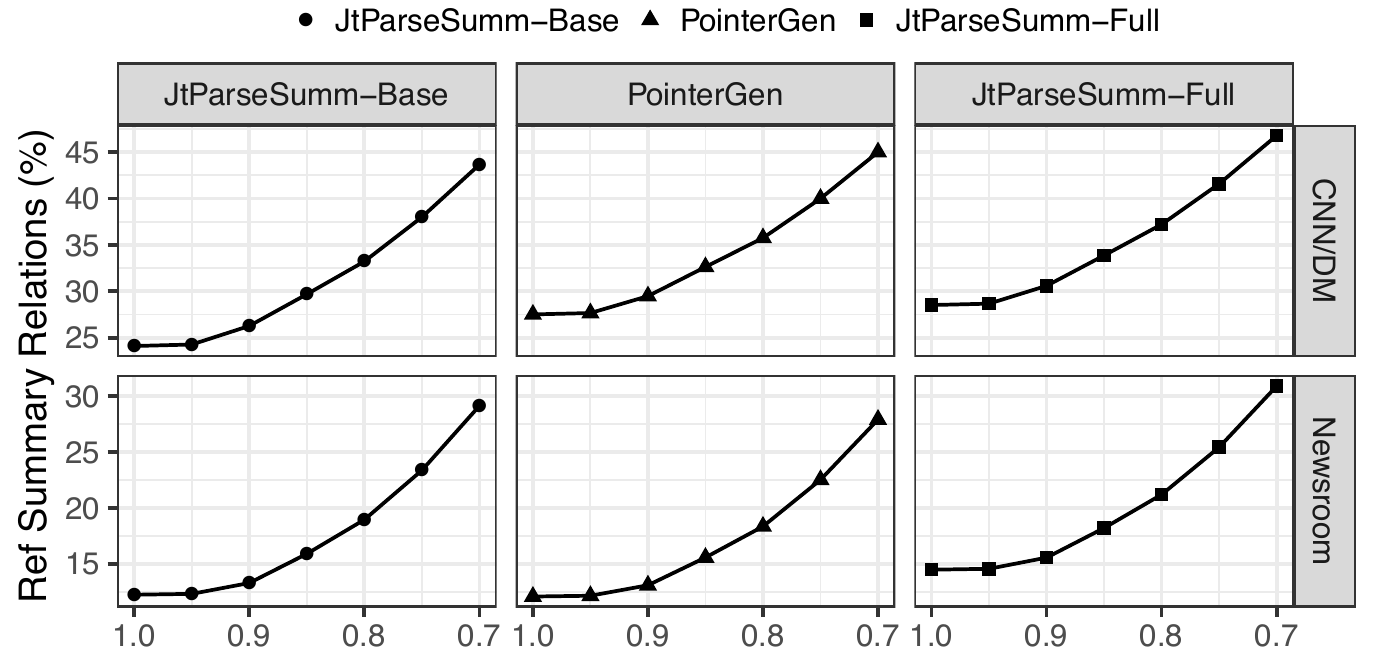}
\includegraphics[width=3in]{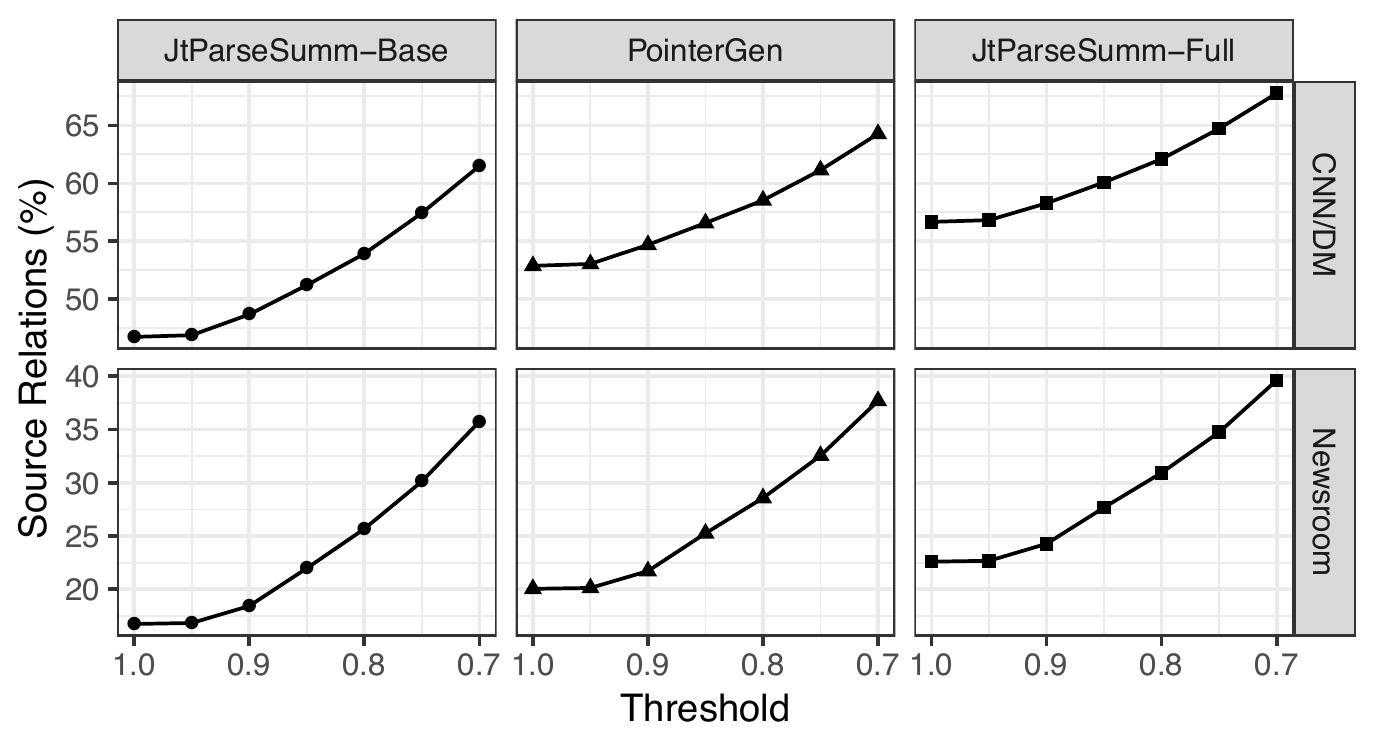}
\caption{F-scores of systems on preserving relations of reference summaries (\emph{top}) and source texts (\emph{bottom}).
We vary the threshold from 1.0 (strict match) to 0.7 in the x-axis to allow for strict and lenient matching of dependency relations.
}
\label{fig:edges}
\end{figure}

\vspace{0.05in}
\noindent\textbf{Dependency parsing}\quad
We expect dependency relations of a summary to be the same or similar to those of the source text or reference summary in order to preserve the original meanings.
Generating a summary word means certain dependency relations are simultaneously added to the summary.
For example, in Table~\ref{tab:rnng}, generating the word \emph{escaped} leads a dependency relation \emph{man}$\leftarrow$\emph{escaped} to be included in the summary.
In this section we demonstrate that by learning to jointly summarize and parse, our system can effectively improve the preservation of dependency relations.\footnote{We cannot compute parsing accuracy, because system and reference summaries use different words and their dependency structures are not directly comparable.}

In Figure~\ref{fig:edges} we demonstrate to what extent system summaries preserve relations of source texts and reference summaries.
We contrast our system GenParse-\textsc{Base} and GenParse-\textsc{Full} that jointly performs summarization and parsing, against the strong baseline of PointerGen that first generates abstractive summaries then parses them using the off-the-shelf Stanford parser~\cite{Chen:2014:EMNLP}.
Dependency relations of source texts and reference summaries are also obtained using the Stanford parser.
We calculate F-scores on preserving reference summary relations (top) and source relations (bottom) and on \textsc{cnn/dm-r} and \textsc{Newsroom} dataset, respectively.
As shown in Figure~\ref{fig:edges}, GenParse-\textsc{Full} consistently outperforms other systems on preserving source and reference summary relations.

Abstractive summaries can contain paraphrases of source descriptions and we thus compare relations using both strict and lenient measures.
A strict measure requires exact match of words.
E.g., two relations $w_{\textsf{1A}}$$\leftarrow$$w_{\textsf{1B}}$ and $w_{\textsf{2A}}$$\leftarrow$$w_{\textsf{2B}}$ are equal if $w_{\textsf{1A}}$ is the same as $w_{\textsf{2A}}$, and $w_{\textsf{1B}}$ is the same as $w_{\textsf{2B}}$.
A lenient measure computes Sim($w_{\textsf{1A}}$,$w_{\textsf{2A}}$) and 
Sim($w_{\textsf{1B}}$,$w_{\textsf{2B}}$) and it requires both scores to be greater than a threshold $\sigma$.
We vary the threshold value along the x-axis to produce the plots in Figure~\ref{fig:edges}.
We define Sim($\cdot$,$\cdot$) as the cosine similarity of word embeddings; and a value of 1.0 corresponds to strict match.
Overall, we notice that the GenParse-\textsc{Full} method performs exceptionally well on retaining relations on the \textsc{cnn/dm-r} dataset.
It achieves an F-score of 56.7\% ($\sigma$=1) / 67.8\% ($\sigma$=0.7) for source relations, and 28.5\% ($\sigma$=1) / 46.8\% ($\sigma$=0.7) for reference summary relations.
This finding suggests that the proposed joint summarization and parsing method performs the best on summaries that contain full grammatical sentences, as is the case with \textsc{cnn/dm-r}, and this matches our expectation.

\vspace{0.05in}
\noindent\textbf{Human evaluation}\quad
We proceed by introducing a novel human evaluation protocol assessing system summaries for grammaticality and preservation of original meanings.
A quantitative measure is important because it allows us to compare different systems regarding to what extent their abstractive summaries preserve the original meanings and whether the summaries contain any falsified content that are nonexistent in the original texts.
The latter is particularly under-investigated in the past.
Our evaluation is made possible by utilizing RDF triples provided in \textsc{WebMerge}.

\begin{table}[t]
\setlength{\tabcolsep}{5pt}
\renewcommand{\arraystretch}{1.05}
\centering
\begin{scriptsize}
\begin{tabular}{ll}
 & \textsf{\emph{\textcolor{red}{Albert B White was born in 1856 and died on July}}} \\
& \textsf{\emph{\textcolor{red}{in Parkersburg, West Virginia.}}}\\
\toprule
\textsf{\textbf{Q1}} & \textsf{How would you rank this summary for grammaticality?}\\
& \textsf{$\square$ 1st (best) \quad\done 2nd \quad\quad$\square$ 3rd \quad$\square$ 4th (worst)}\\[0.3em]
\textsf{\textbf{Q2a}} & \textsf{Does this summary convey the following meaning?}\\
& \textsf{\emph{\textcolor{blue}{(Albert B. White $|$ birthYear $|$ 1856)}}} \\
& \textsf{\done Yes \quad$\square$ No} \\[0.3em]
\textsf{\textbf{Q2b}} & \textsf{Does this summary convey the following meaning?}\\
& \textsf{\emph{\textcolor{blue}{(Albert B. White $|$ deathPlace $|$ Parkersburg, West Virginia)}}}\\
& \textsf{\done Yes  \quad$\square$ No}\\[0.3em]
\textsf{\textbf{Q3}} & \textsf{Does this summary convey any \textcolor{blue}{additional meanings} not}\\
& \textsf{covered by the above triples?}\\
& \textsf{\done Yes \quad$\square$ No}\\
\bottomrule
\end{tabular}
\end{scriptsize}
\caption{We present a summary to a group of human judges.
They are instructed to assess the summary for grammaticality and preservation of original meanings.}
\label{tab:mturk}
\end{table}

Table~\ref{tab:mturk} illustrates the evaluation process.
We present a summary to a group of human judges.
They are instructed to rank this summary among four peers for grammaticality.
Next, we require the judges to answer a set of binary questions on (\textsf{\footnotesize Q2}) if the summary has conveyed the meaning of a given RDF triple, and (\textsf{\footnotesize Q3}) if the summary has conveyed any additional meanings that are not in the collection of triples.
In particular, an RDF (Resource Description Format) triple is of the form \emph{subject $|$ property $|$ object} and it is used for meaning representation~\cite{Narayan:2017:EMNLP}.
The number of triples per instance varies from 1 to 7.
A successful summary should preserve the meanings of all RDF triples and it shall not introduce any additional meanings.
As an example, the summary \textsf{A} in Table~\ref{tab:mturk} has introduced undesired content during abstraction (\emph{died on July}), it thus makes factual errors that can mislead the reader.

Peer summaries are generated by GenParse-\textsc{Base}, PointerGen, and GenParse-\textsc{Full}.
We further include human summaries for comparison; the order of presentation of summaries is randomized.
We sample 100 instances from the test set of \textsc{WebMerge} and employ 5 human judges on Amazon mechanical turk (mturk.com) to perform the task; they are rewarded \$0.1 per HIT.
Importantly, we are able to filter out low-quality responses from AMT judges using their answers for human summaries,
as they are expected to answer unanimously \emph{yes} for \textsf{\footnotesize Q2} and \emph{no} for \textsf{\footnotesize Q3}.
We expect this method to improve the quality and objectivity of human evaluation.

We present evaluation results in Table~\ref{tab:results_human}.
It is not surprising that human summaries are ranked 1st on grammaticality.
Our GenParse-\textsc{Full} method consistently outperforms its counterparts and it is ranked 2nd best followed by PointerGen and GenParse-\textsc{Base}.\footnote{We perform pairwise comparisons between systems.
Results reveal that there is no significant difference between GenParse-\textsc{Base} and PointerGen.
All other differences are statistically significant according to a one-way ANOVA with posthoc Tukey HSD test ($p<$0.01).}
We report the system accuracy on preserving source semantics (\textsf{\footnotesize Q2}) and preventing system summaries from changing original meanings (\textsf{\footnotesize $\boldsymbol\neg$Q3}). 
Our method (GenParse-\textsc{Full}) again excels in both cases.
But the scores (46.8 and 53.4) suggest that ensuring abstractive summaries to preserve source content remains a challenging task, and similar findings are revealed by Cao et al.~\shortcite{Cao:2018} and See et al.~\shortcite{See:2017}.
Our results are highly encouraging.
The human evaluation protocol is particularly meaningful to quantitatively measure to what extent system-generated abstractive summaries remain true-to-original.

\begin{table}[t]
\setlength{\tabcolsep}{3.5pt}
\renewcommand{\arraystretch}{1.08}
\centering
\begin{small}
\begin{tabular}{l|rrrr|rr}
\multicolumn{1}{c}{} & \multicolumn{4}{c}{\textbf{Grammaticality}} & \multicolumn{2}{c}{\textbf{Meaning}}\\
& \multicolumn{1}{c}{\textbf{1st}} & \multicolumn{1}{c}{\textbf{2nd}} & \multicolumn{1}{c}{\textbf{3rd}} & \multicolumn{1}{c|}{\textbf{4th}} & \multicolumn{1}{c}{\textbf{Q2}} & \multicolumn{1}{c}{$\boldsymbol\neg$\textbf{Q3}}\\
\toprule
Human & \textbf{73.7} & 15.3 & 5.9 & 5.1 & 100 & 100\\
GenParse-\textsc{Base} & 8.5 & 23.7 & 30.5 & 37.3 & 15.8 & 12.7\\
PointerGen & 5.1 & 18.6 & \textbf{35.6} & \textbf{40.7} & 38.5 & 50.8\\
GenParse-\textsc{Full} & 12.7 & \textbf{42.4} & 28.0 & 17.0 & \textbf{46.8} & \textbf{53.4}\\
\bottomrule
\end{tabular}
\end{small}
\caption{Human assessment of grammaticality and semantic accuracy of various summaries.
Our GenParse-\textsc{Full} achieves the best results on both aspects among all systems.
}
\label{tab:results_human}
\end{table}

\section{Conclusion}
\label{sec:conclusion}

We propose to jointly summarize and parse to improve the grammaticality and truthfulness of summaries.
We introduce a neural model combining a sequential decoder with a tree-based decoder and ensure both work in a synchronized manner. 
Experimental results show that our method performs on par with or superior to state-of-the-art systems on standard test sets. 
It surpasses strong baselines on human evaluation of grammaticality and preservation of meanings.

\section*{Acknowledgments}
We are grateful to the reviewers for their valuable comments and suggestions. 
This research was supported in part by the National Science Foundation grant IIS-1909603.

\begin{small}
\bibliography{parsing,fei,summ,abs_summ}
\bibliographystyle{aaai}
\end{small}

\end{document}



\appendix

\begin{table*}[t]
\setlength{\tabcolsep}{4pt}
\renewcommand{\arraystretch}{1.15}
\centering
\begin{footnotesize}
\begin{tabular}{|c|r|p{5in}|}
\hline
\multirow{5}{*}{\rotatebox[origin=c]{90}{\textsf{Gigaword}}} & \textbf{Source Text} & \emph{A man who pushed his wife to her death from a sixth-floor balcony has been executed , the China Women 's News reported Friday}\\
& \textbf{Human} & Man Who Pushed Wife Off Balcony , Others Executed in China \\
& \textbf{GenParse-\textsc{Base}} & china executes man who pushed wife to death \\
& \textbf{PointerGen} & man executed for pushed wife to balcony  \\
& \textbf{GenParse-\textsc{Full}} & china executes man who pushed wife to her death \\
\hline
\hline
\multirow{5}{*}{\rotatebox[origin=c]{90}{\textsf{Gigaword}}} & \textbf{Source Text} & \emph{Vietnam has recently detected a small number of chickens being infected with bird flu virus strain of H5 , and remained highly alert for potential new outbreaks in winter when weather conditions favor the development of viruses .} \\
& \textbf{Human} & Vietnam detects new bird flu infections  \\
& \textbf{GenParse-\textsc{Base}} & vietnam detected chickens infected with bird flu\\
& \textbf{PointerGen} & new bird flu virus detected in vietnam \\
& \textbf{GenParse-\textsc{Full}} & vietnam detected chickens infected with bird flu virus \\
\hline
\hline
\hline
\multirow{5}{*}{\rotatebox[origin=c]{90}{\textsf{Gigaword}}} & \textbf{Source Text} & \emph{Around 300 clandestine immigrants Wednesday staged a peaceful breakout from a detention centre in Malta and demonstrated on a road shouting `` We want freedom . ''} \\
& \textbf{Human} & Mass breakout of immigrants from Malta centre \\
& \textbf{GenParse-\textsc{Base}} & illegal immigrants stage peaceful life in malta \\
& \textbf{PointerGen} & 300 illegals demonstrate in malta \\
& \textbf{GenParse-\textsc{Full}} & clandestine immigrants stage peaceful breakout in malta \\
\hline
\hline
\multirow{5}{*}{\rotatebox[origin=c]{90}{\textsf{Gigaword}}} & \textbf{Source Text} & \emph{In a twist of the U.S.-Russian controversy over Iran 's nuclear program , Russia 's atomic energy minister on Friday suggested Washington join Moscow in building a nuclear power plant in Iran .} \\
& \textbf{Human} & Russian minister offers United States to join Moscow 's nuclear deal \\
& \textbf{GenParse-\textsc{Base}} & russia suggests u.s. join moscow in building nuclear power plant\\
& \textbf{PointerGen} & russian atomic energy minister suggests washington join moscow in \\
& \textbf{GenParse-\textsc{Full}} & russian atomic energy minister suggests u.s. join moscow nuclear power plant in iran\\
\hline
\hline
\multirow{5}{*}{\rotatebox[origin=c]{90}{\textsf{Newsroom}}} & \textbf{Source Text} & \emph{A suspended Melbourne pharmacist allegedly stole more than \$ 80,000 of drugs from a company linked to the Essendon AFL club doping scandal .}\\
& \textbf{Human} & Vic pharmacist allegedly stole peptides \\
& \textbf{GenParse-\textsc{Base}} & melbourne pharmacist allegedly stole more than \$ 80,000 \\
& \textbf{PointerGen} & melbourne pharmacist allegedly stole more than \$ 80,000 \\
& \textbf{GenParse-\textsc{Full}} & melbourne pharmacist allegedly stole more than \$ 80,000 of drugs from doping scandal \\
\hline
\hline
\multirow{7}{*}{\rotatebox[origin=c]{90}{\textsf{CNN/DM}}} & \textbf{Source Text} & \emph{The 40-year-old rapper from St. Louis , who shot to fame 15 years ago with the track `` Country Grammar , '' has been charged with felony possession of drugs , simple possession of marijuana and possession of drug paraphernalia , the Tennessee Department of Safety and Homeland Security said.} \\
& \textbf{Human} &  nelly has been charged with felony possession of drugs\\
& \textbf{GenParse-\textsc{Base}} & the 40-year-old has been charged with felony possession of drugs\\
& \textbf{PointerGen} & he shot to fame 15 years ago with the track `` country grammar `` \\
& \textbf{GenParse-\textsc{Full}} & he is charged with felony possession of drugs , simple possession of drug paraphernalia\\
\hline
\hline
\multirow{5}{*}{\rotatebox[origin=c]{90}{\textsf{WebMerge}}} & \textbf{Source Text} & \emph{The icebreaker , Aleksey Chirikov has a ship beam of 21.2 m. The icebreaker Aleksey Chirikov was built in Helsinki .}\\
& \textbf{Human} & The icebreaker Aleksey Chirikov was built in Helsinki and has a ship beam of 21.2 m. \\
& \textbf{GenParse-\textsc{Base}} & the aidaluna has a beam of . metres and a beam of . m.\\
& \textbf{PointerGen} & built in helsinki the beam of aleksey and the icebreaker of the icebreaker beam\\
& \textbf{GenParse-\textsc{Full}} & the icebreaker aleksey chirikov chirikov was built in helsinki and has a ship beam of 21.2 m.\\
\hline
\end{tabular}
\end{footnotesize}
\caption{We present example summaries generated by various systems.
Our GenParse system performs on par with or superior to state-of-the-art systems on standard test sets. 
The full model outperforms the GenParse-\textsc{Base} model, demonstrating the effectiveness of combining a sequential decoder with tree-based decoder in a synchronized manner.}
\label{tab:system_output}
\end{table*}